\documentclass{czipreprint}

\usepackage{natbib}
\setcitestyle{authoryear,open={(},close={)}} %Citation-related commands

\usepackage[usenames,dvipsnames]{xcolor}

\hypersetup{
    colorlinks=true,
    linkcolor=black,
    urlcolor=blue,
    citecolor=gray,
    }

\usepackage{MyPackages}

\usepackage{MyDefinitions}

\graphicspath{ {./img/}{./} }

\organization{CZI} % The default

\begin{document}

\title[How Well Do LLMs Understand Drug Mechanisms?]{How Well Do LLMs Understand Drug Mechanisms? A Knowledge + Reasoning Evaluation Dataset}

% Use optional arguments to show affiliations.  Starred command is for
% corresponding authors
\author*[1]{Sunil Mohan}
\email{smohan@chanzuckerberg.com}

\author[1]{Theofanis Karaletsos}
\email{tkaraletsos@chanzuckerberg.com}
\orcid{0000-0002-0296-3092}

%\affiliation[1]{Chan Zuckerberg Initiative, Redwood City, CA}

% Put the date in the format YYYY-MM-DD or \today
\date{\today}
%\date{2024-10-04}

% Use CZI DOI *only* if instructed by Comms!
% ArXiv, bioRxiv, medRxiv assign their own DOI numbers
%\preprintdoi{10.1111/XXX.XXXX.YYY}

\maketitle

\begin{center}
$^{\text{a}}$ Chan Zuckerberg Initative, Redwood City, CA\\
\end{center}
\vspace{0.2in}

\begin{abstract}
Two scientific fields showing increasing interest in pre-trained large language models (LLMs) are drug development / repurposing, and personalized medicine. For both, LLMs have to demonstrate factual knowledge as well as a deep understanding of drug mechanisms, so they can recall and reason about relevant knowledge in novel situations. Drug mechanisms of action are described as a series of interactions between biomedical entities, which interlink into one or more chains directed from the drug to the targeted disease. Composing the effects of the interactions in a candidate chain leads to an inference about whether the drug might be useful or not for that disease. We introduce a dataset that evaluates LLMs on both factual knowledge of known mechanisms, and their ability to reason about them under novel situations, presented as counterfactuals that the models are unlikely to have seen during training. Using this dataset, we show that o4-mini outperforms the 4o, o3, and o3-mini models from OpenAI, and the recent small Qwen3-4B-thinking model closely matches o4-mini's performance, even outperforming it in some cases. We demonstrate that the open world setting for reasoning tasks, which requires the model to recall relevant knowledge, is more challenging than the closed world setting where the needed factual knowledge is provided. We also show that counterfactuals affecting internal links in the reasoning chain present a much harder task than those affecting a link from the drug mentioned in the prompt.\footnote{This is a slightly expanded version of our paper \textit{How Well Does ChatGPT Understand Drug Mechanisms? A Knowledge + Reasoning Evaluation Dataset}, published in FLLM 2025 \citep{Mohan-Karaletsos:2025}.}
\end{abstract}

%\tableofcontents
%
%\clearpage
%\listoffigures
%\listoftables
%\listofalgorithms
%\lstlistoflistings
%
%\clearpage

%%==============================================================================================

%
% Put sections into separate files
%

\section{Introduction}
\label{sec:Intro}

Scientific research requires analysing novel situations based on deep knowledge and understanding of the domain.  For general purpose pre-trained large language models (LLMs) to be useful tools in such endeavors, they need to be able to bring both factual knowledge and a domain-specific reasoning ability to the tasks. There have been several efforts to investigate and chart the reasoning ability of LLMs, both from a task-specific perspective as well as understanding the core capabilities \citep{Sun-etal:2024,Yu-etal:2024,Mondorf-etal:2024}. Researchers have observed that LLMs tend to perform better at recalling factual information than on reasoning (e.g. \citep{Huyuk-etal:2025}). Researchers have also recognized the importance of using counterfactuals (e.g. \citep{Wu-etal:2024}) and other variants of known problems \citep{Srivastava-etal:2024}, which describe scenarios the model is unlikely to have seen during training, for obtaining more robust and generalizable metrics of LLM reasoning.

As LLMs show improving capabilities, they are increasingly being investigated as tools in various research and clinical tasks in medicine, e.g. clinical decision support \citep{Benary-etal:2023}, personalized medicine \citep{Gao-etal:2025:TxAgent}, and drug development and repurposing \citep{AbuNasser-etal:2024,Gottweis-etal:2025:Google}. We focus on these fields where, for LLMs to be useful, they need to be able to recall factual knowledge about the interactions of drugs and diseases in the body, and also exhibit a deep enough understanding of the language these are described in to be able to reason about them, especially under novel situations. We developed the Drug Mechanisms Counterfactuals Dataset to investigate this capability.

DrugBank defines a Mechanism of Action\footnote{\url{https://dev.drugbank.com/guides/terms/mechanism-of-action}} (MoA) as:
\begin{quote}
``the specific chemical and/or physical effects, activities, processes, etc. that a particular drug either elicits or participates in so as to fulfill one or more medical indications.''
\end{quote}
In pharmaceutical databases like DrugBank, MoAs are presented as descriptive text as part of that drug's information\footnote{\url{https://go.drugbank.com/drugs/DB0104\#mechanism-of-action} describes the MoA for the drug Abacavir in DrugBank.}.
These describe a series of interactions between biomedical entities like chemicals, genes, proteins and biological processes. Taken together, they form one or more interlinked chains of effects and relationships that counteract the corresponding effects or causes (if the cause is internal) of the disease that drug is an indication for (see fig.~\ref{fig:MoA_example} for an example).  One can take any such set of interactions, check to see if it forms an interlinked path between a drug and a disease, compose them into an inference chain, and reason about whether that chain forms a potential MoA where the drug affects the disease in a therapeutic manner, or exacerbates it.

This is the representation that the Drug Mechanisms Counterfactuals Dataset is based on. The counterfactuals are derived by altering links in such inference chains, or initiating new ones, and the LLM is asked about the effect of that change. In the rest of the paper, we introduce the dataset, and use it to evaluate some recent OpenAI LLMs. We show that the reasoning models o3, o3-mini and o4-mini outperform GPT-4o at these tasks, with o4-mini showing the best results despite being a smaller model than o3. We test our queries in both the closed world setting, where the needed factual interactions and MoA knowledge is included in the prompt, and the open world setting where the model is expected to recall all relevant knowledge for the task. Results show that open world is the more challenging setting. Our evaluation also reveals that `deep' counterfactuals that are based on internal links in the MoA inference chain present a much harder problem than `surface' counterfactuals that are based on a direct interaction with the drug mentioned in the prompt.

This paper is a slightly expanded version of our FLLM 2025 paper \citep{Mohan-Karaletsos:2025}, in that it includes some additional figures, and also an evaluation of the Qwen3-4B Thinking LLM.

The Drug Mechanisms Counterfactuals Dataset and evaluation framework is available for download\footnote{\url{https://github.com/czi-ai/DrugMechCounterfactuals}}.

\section{Drug Mechanisms and Counterfactuals}
\label{sec:Dataset}

% --------------------------------------------------------------------------------------------------------------------------------------
\subsection{Drug Mechanisms of Action (MoA)}

\begin{figure}
\centering
\includegraphics[width=.8\columnwidth]{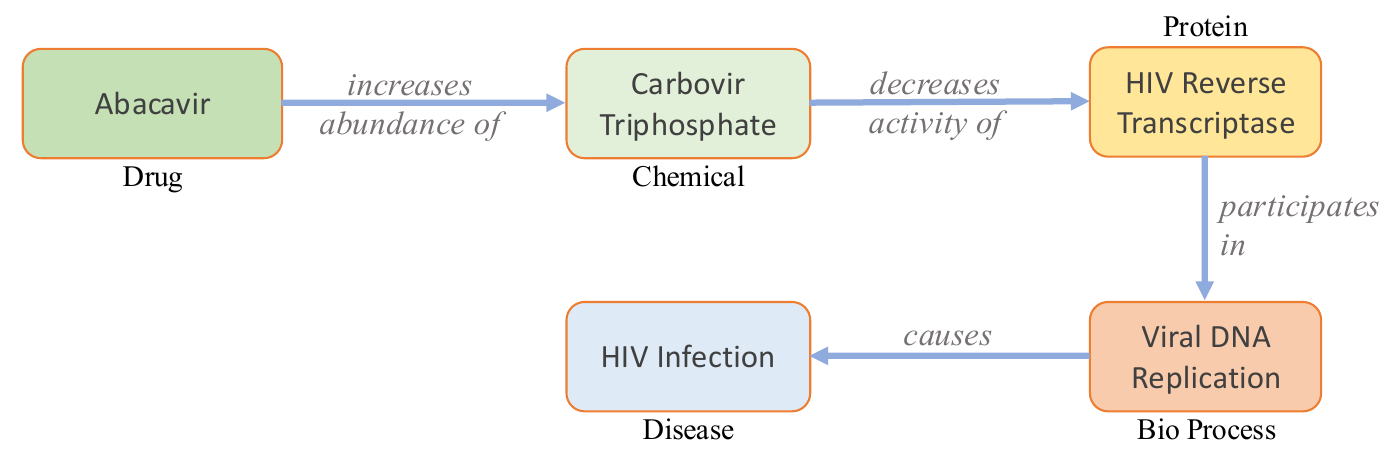}

{\tiny
\begin{verbatim}
Drug:Abacavir  | increases abundance of |  Chemical:carbovir triphosphate
Chemical:carbovir triphosphate  | decreases activity of \
    |  Protein:HIV-1 reverse transcriptase
Protein:HIV-1 reverse transcriptase  | participates in  \
    |  BiologicalProcess:Viral DNA replication
BiologicalProcess:Viral DNA replication  | causes |  Disease:HIV infection
\end{verbatim}
}
\caption{Two representations of the MoA for Abacavir on HIV Infection.}
\label{fig:MoA_example}
\end{figure}

The Drug Mechanisms Counterfactuals dataset is derived from DrugMechDB \citep{DrugMechDB:2023}, a database of graphical representations of the Mechanisms of Action of 4,664 Drug-Disease pairs, manually curated ``from free-text descriptions found on DrugBank, Wikipedia and within literature''.  The curators encoded 14 entity types following specific ontologies\footnote{\url{https://sulab.github.io/DrugMechDB\#curation}}, and 66 interaction relation (edge) labels, which don't have formal definitions. Each interaction is directed from a source to a target entity. Its label describes how the source relates to (e.g. ``located in'') or affects (e.g. direction-of-effect like ``increases abundance of'', ``negatively regulates'') the target entity.

DrugMechDB curators followed the convention that each MoA is a directed acyclic graph (DAG), and the Drug entity serves as the single root-node (no incoming edges) and the Disease entity as the single sink-node (no outgoing edges), such that all paths between the two are directed from the Drug to the Disease. An example is shown in fig.~\ref{fig:MoA_example}.

% --------------------------------------------------------------------------------------------------------------------------------------
\subsection{Evaluating Consistency between two MoA Graphs}
\label{sec:MoA-Consistency}

Some of our experiments ask ChatGPT to respond with a MoA graph, using the textual format shown in fig.~\ref{fig:MoA_example}. While this makes matching it to a reference MoA easier, there is no guarantee that ChatGPT will follow the same formal names for entities and interactions. Furthermore, it is common in medical literature to skip some interactions, or group a series of molecular interactions into pathways or biological processes. So instead of looking for an exact graph match, we test for directional consistency, as described below.

Entity names (nodes) are matched using approximate text match; the prompt asks for a specific drug and disease, and those are mapped directly to the response. The candidate DAG $G_c = (N_c, E_c)$ is defined to be \textit{directionally consistent} with the reference DAG $G_r = (N_r, E_r)$ under the partial node mapping $f$ if, for every simple directed path from $n_1$ to $n_2$ in $G_r$, where $n_1, n_2 \in N^f_r$, there is a directed path from $f(n_1)$ to $f(n_2)$ in $G_c$. Note that this consistency is non-symmetric, and though edges are directed, their labels are ignored.

Using this definition, we can compute the proportion of \textit{interior nodes} (not including the drug and disease) that matched, and the proportion of reduced edges in $G_c$ that were directionally consistent with $G_r$ (ideally all). The MoA returned by ChatGPT is considered `potentially very different' if none of the interior nodes matched.

% --------------------------------------------------------------------------------------------------------------------------------------
\subsection{Dataset for Factual Knowledge of MoAs}
\label{sec:Data-MoAs}

The first step is to establish whether ChatGPT knows about well known drug MoAs. For this, we used 1,000 Drug-Disease MoAs randomly drawn from DrugMechDB. Negative samples were constructed from PrimeKG \citep{Chandak-etal:2023:PrimeKG}, with 500 randomly drawn Drug-Disease pairs with a \textit{contra-indication} relationship, and another 500 pairs for which there was no direct Drug-Disease relationship in PrimeKG. For positive samples, the expected response is a (matching) MoA. For negative samples, the LLM should not be able to find a valid MoA.

% --------------------------------------------------------------------------------------------------------------------------------------
\subsection{The Drug Mechanisms Counterfactuals Dataset}
\label{sec:Data-Counterfactuals}

The Counterfactuals dataset is designed to test an LLM's understanding of MoA's by asking it to reason about novel situations presented as counterfactuals. Each counterfactual involves a new relationship between a pair of entities. The query to the LLM presents it as an observation happening in the presence of an experimental drug being administered to patients exhibiting an unusual combination of symptoms. Drugs are known to change interactions in the body, so this does not violate the biology model an LLM may have learnt.

\begin{figure}
\centering
\includegraphics[width=\columnwidth]{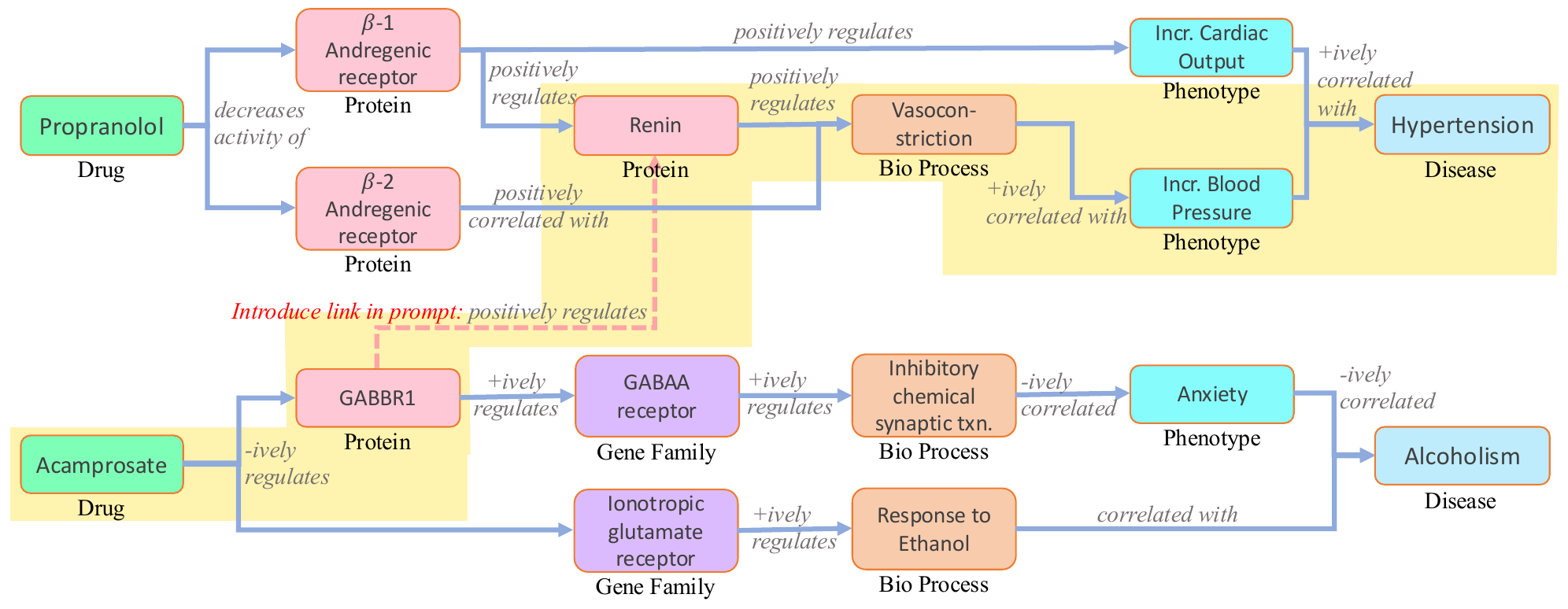}
\caption[Synthesizing a new MoA by adding an interaction]{Synthesizing a new MoA (highligted) by adding an interaction (red dashed arrow). With the synthesized MoA, the drug Acamprosate is now effective in treating Hypertension.}
\label{fig:AddLink_example}
\end{figure}

For each counterfactual in the Drug Mechanisms Counterfactuals Dataset, two entities are chosen from DrugMechDB, and their interaction status is changed to cause a specific type of effect on drug mechanisms:
\begin{enumerate}
\itemi{Add-Link}: In positive samples, a new interaction is introduced between two entities, chosen from two MoAs for different drug-disease pairs, such that a new MoA is now enabled for \textit{drug-1} from the first MoA treating \textit{disease-2} from the second (see fig.~\ref{fig:AddLink_example}). In negative samples, the direction-of-effect of the new interaction is inverted  (e.g. ``increases abundance of'' now becomes ``decreases abundance of'', ``negatively regulates'' becomes ``positively regulates''). As a result, even though there is a chain of interactions connecting the drug to the disease, they do not compose into a therapeutic action on the disease  (does not help the disease and may exacerbate it). The model is then asked whether, for those patients, \textit{drug-1} might treat \textit{disease-2}, and to describe the MoA.

\itemi{Delete-Link}: In positive samples, an interaction in an existing MoA is rendered inactive. The interaction is chosen as a cut-edge in the MoA graph, so its `deletion' means the MoA's \textit{drug} is no longer connected to the \textit{disease}. In negative samples, the deleted interaction is from an entity in the MoA to some random protein not participating in that MoA. The LLM is asked whether that MoA's \textit{drug} might be useful in treating that \textit{disease} in the patients where this counterfactual was observed.

\itemi{Invert-Link}: In positive samples, a cut-edge interaction is chosen from an existing MoA, as for Delete-Link, and its direction-of-effect is inverted. This renders that MoA therapeutically invalid. In negative samples, the observed interaction is from the source entity to a randomly chosen protein not participating in the chosen MoA. The LLM query is the same as for Delete-Link.
\end{enumerate}
The negative samples are chosen to check whether the LLM is simply following some syntactic cues, or actually reasoning about the task.
These categories have two further sub-types based on the location of the source entity in the source MoA: (i) it is the drug, or (ii) it is an interior node in that MoA. Viewing the MoA as an inference chain from the drug to the disease, the first case involves a `\textit{surface}' link in that chain, where the source entity of the counterfactual interaction is the drug mentioned in the prompt. The second case presents a `\textit{deep}' link, where the relationship of the counterfactual interaction to the drug is not explicit in the prompt. In all cases, there is no direct interaction with the prompt's disease.

The Add-Link example in fig.~\ref{fig:AddLink_example} involves a `\textit{deep counterfactual}': the source node is an interior protein node (`GABBR1'), at a distance 1 edge away from the corresponding drug (`Acamprosate'). The resulting synthesized MoA, highlighted in yellow, is 5 edges long.

The \textit{Add Link} set has 1,000 samples for each of the two sub-types, for both positive and negative. \textit{Delete} and \textit{Invert} counterfactuals have 250 samples for each of the eight cases.

The MoAs and the participating entities are chosen at random, subject to the constraints described above, to further ensure that the counterfactual situations being described are unlikely to have been encountered in the model's training data. We developed a mechanism of action simulator that checked the validity of the inference chains represented by paths in a drug-disease MoA. This simulator was used to validate the interaction labels
on the Add and Invert-Link counterfactuals.

The desired response for positive Add-Link samples is a MoA, and for negatives, that no supporting MoA could be found. For Delete and Invert Link, the LLM is presented with four optional answers, shown with convenience labels below:
\begin{itemize}
\item {No Effect}: ``\textit{The observed behavior will not affect the drug's mechanism of action on the disease. The drug's intended effect is still active, so the drug can be used to treat the disease in these patients.}''

\item {Partially Blocked}: ``\textit{The observed behavior affects only one of the drug's mechanisms of action. However the drug has other mechanisms that are unaffected, so the drug could still provide therapeutic benefits in treating the disease in these patients.}''

\item {Fully Blocked}: ``\textit{The observed behavior suggests that the drug's mechanism of action is completely blocked in these patients. So the drug will not be effective in treating the disease in these patients.}''

\item {Contra-indicated}: ``\textit{Given the observed behavior, the drug could worsen the disease in these patients and potentially exacerbate its symptoms. So the drug should not be used on these patients.}''
\end{itemize}
The desired responses are:
\begin{itemize}
\itemi{Delete-Link}. Positive samples: the MoA should become \textit{Fully Blocked}; Negative: the correct option is \textit{No Effect}.

\itemi{Invert-Link}. Positive: the MoA is \textit{Contra-indicated} or \textit{Fully Blocked}; Negative: the correct option is \textit{No Effect}.
\end{itemize}

\section{Experiments}
\label{sec:Experiments}

We used the Drug Mechanisms Counterfactuals Dataset to evaluate four OpenAI models\footnote{\url{https://platform.openai.com/docs/models}}, and one Qwen model:
\begin{itemize}
\itemi{GPT-4o}: listed as their flagship ``fast, intelligent and flexible'' model. Version {gpt-4o-2024-08-06}, with a knowledge cutoff date Sep 30, 2023.

\itemi{o3}: Their ``most powerful reasoning model''; ver. {o3-2025-04-16}, knowledge cutoff May 31, 2024.

\itemi{o3-mini}: Smaller version of o3; ver. {o3-mini-2025-01-31}, knowledge cutoff May 31, 2024.

\itemi{o4-mini}: Latest small o-series; ver. {o4-mini-2025-04-16}, knowledge cutoff May 31, 2024.

\itemi{Qwen3-4B}: A recent small `thinking' version of Qwen3 designed for complex reasoning tasks, ver. {Qwen3-4B-Thinking-2507-FP8}\footnote{\url{https://huggingface.co/Qwen/Qwen3-4B-Thinking-2507-FP8}}.
\end{itemize}
Each test sample was issued as a separate query in a separate session through the OpenAI API. The following API parameters were used for all calls: seed = 42, temeprature = 1.0, presence\_penalty = 0, max\_completion\_tokens = None, and for the OpenAI reasoning models, reasoning\_effort = ``medium''. Metrics for the counterfactual queries reported below are averaged across five runs.

The Qwen model was deployed locally using a single \texttt{vllm}-based server\footnote{\url{https://docs.vllm.ai/en/latest/serving/openai_compatible_server.html}}, and given a timeout of 600 seconds. Due to time constraints, only a single run was performed for this model.

% --------------------------------------------------------------------------------------------------------------------------------------
\subsection{Factual Knowledge of MoAs}
\label{sec:Exp-KnownMoAs}

These experiments were used to establish how well ChatGPT would respond to requests for the mechanisms of action of known and approved drugs. We only tested the GPT-4o model (temperature = 0.2) for this, and also used this as a test for our MoA consistency metrics (\S~\ref{sec:MoA-Consistency}).

We devised three query modes that varied on what information was presented about the disease in the prompt (all three modes explicitly named the drug):
\begin{enumerate}
\itemi{Named Disease}: The disease is explicitly named.

\itemi{Anonymized Disease}: The disease is not named, presented instead as ``\textit{a new disease}'', and described via its directly associated entities in the MoA in DrugMechDB. This mode is closer to evaluations done for Machine Learning models using held out data. A recent study even observed a performance drop in popular LLMs when the generic drug name in the query was replaced with the corresponding brand name \citep{Gallifant-etal:2024}. In our reference data and our experiments, the average number of entities directly associated with the disease is 1.2.

We tested providing disease characteristics taken from the description in MONDO. However ChatGPT recognized the disease, which defeated the test's purpose.

\itemi{Named Disease with Associations}: To complete the spectrum of evaluations, this mode tests whether providing both the name of the Disease as well as its directly associated entities affected the model's response.
\end{enumerate}
The results for these modes are shown in Table~\ref{tab:Known-MoA}, mean $\pm$ s.d. (rounded to 2 digits) averaged across 5 runs. Accuracy for positives means a MoA was returned, and for negatives that no MoA was returned.

\begin{comment}
%
% NOTE: This table replaced with expanded version below
%
\begin{table}
\renewcommand{\arraystretch}{1.3}
\centering
\caption{Experiment results for Factual Knowledge of MoAs.}
%\label{tab:Known-MoA}
\begin{tabular}{lr}
\hline
\bfseries Metric & \bfseries GPT-4o\\
\hline
Accuracy (average)  &  $0.92$                  \\
\ldots on $+$ives   &  $0.85 \pm 0.00$   \\
\ldots on $-$ives    &  $0.98 \pm 0.00$   \\
Potentially v. different MoAs       &     $8\% \pm 0\%$  \\
avg. Interior Node-match score   &    $0.37 \pm 0.00$  \\
avg. Reduced Edge-match score  &    $0.90 \pm 0.00$  \\
\hline
\end{tabular}
\end{table}
\end{comment}

\begin{table}
\renewcommand{\arraystretch}{1.3}
\centering
\caption{Factual Knowledge of MoAs: Measuring the consistency of the returned MoA to the reference, model GPT-4o.}
\label{tab:Known-MoA}
%\resizebox{\columnwidth}{!}{%
\begin{tabular}{lrrr}
\hline
\bfseries Metric  & \bfseries Anonym. & \bfseries Named & \bfseries Name + Assoc.\\
\hline
Accuracy (average)  &   $0.93$                  &  $0.92$                  &   $0.94$ \\
\ldots on $+$ives   &    $0.89 \pm 0.01$   &  $0.85 \pm 0.00$   &   $0.90 \pm 0.00$ \\
\ldots on $-$ives    &    $0.97 \pm 0.00$   &  $0.98 \pm 0.00$   &   $0.97 \pm 0.00$ \\
%Pot. very diff. MoAs       & $26\% \pm 0\%$    &     $8\% \pm 0\%$  &   $24\% \pm 0\%$   \\
Interior Node-match   & $0.50 \pm 0.00$    &    $0.37 \pm 0.00$  &    $0.51 \pm 0.00$   \\
Reduced Edge-match  & $0.98 \pm 0.00$    &    $0.90 \pm 0.00$  &    $0.98 \pm 0.00$   \\
\hline
\end{tabular}
%}
\end{table}

A primary issue was getting GPT-4o to respond with a well structured MoA that also matched the reference, on entity types and entity names. Using in-context learning \citep{Brown-etal:2020} was key, in addition to explicit instructions: we used 4 positive examples and one negative in the prompt (distinct from the test samples). The general format of the prompt followed that used in \citep{Edge-etal:2024:GraphRag}. Even then, GPT-4o's response would sometimes be mis-structured, which affected the graph consistency metrics.

Interestingly, the model performed better on negative samples. For positive samples, when an MoA is returned, on average a third of the interior nodes (not the drug or disease) matched to the reference in the \textit{Named Disease} mode, indicating different expressions of the same MoA. Interior node match for the \textit{Anonymized} and \textit{Named with Associations} modes do not count the associated entities mentioned in the prompt. Despite that, these scores are higher. We believe this is because the associated entities mentioned in the prompt provide an indication to the LLM which version of the MoA is being requested, from among the multiple representations it encountered during training. The reduced edge match scores are quite high, indicating that most of the time the returned MoA was directionally consistent with the reference.

% --------------------------------------------------------------------------------------------------------------------------------------
\subsection{Reasoning about Drug Mechanisms with Counterfactuals}
\label{sec:Exp-Counterfactuals}

All five Large Language models were tested on the reasoning tasks, using the parameters described above, but only the \textit{Named Disease} mode. Repeating the runs for the OpenAI models yielded very low variability in the responses (s.d. $\le 0.03$). Add-Link queries require the LLM to respond with an MoA. As for the factual knowledge test (\S~\ref{sec:Exp-KnownMoAs}), we used examples in the prompt: 5 positive and 3 negative. Queries for Delete and Edit-Link were zero-shot, asking the LLM to explain its reasoning and then select the best answer option (see \S~\ref{sec:Data-Counterfactuals}). For the closed world setting, the relevant MoAs were inserted into the same prompt, including for the examples in Add-Link.

\begin{figure}
\centering
\includegraphics[width=\columnwidth]{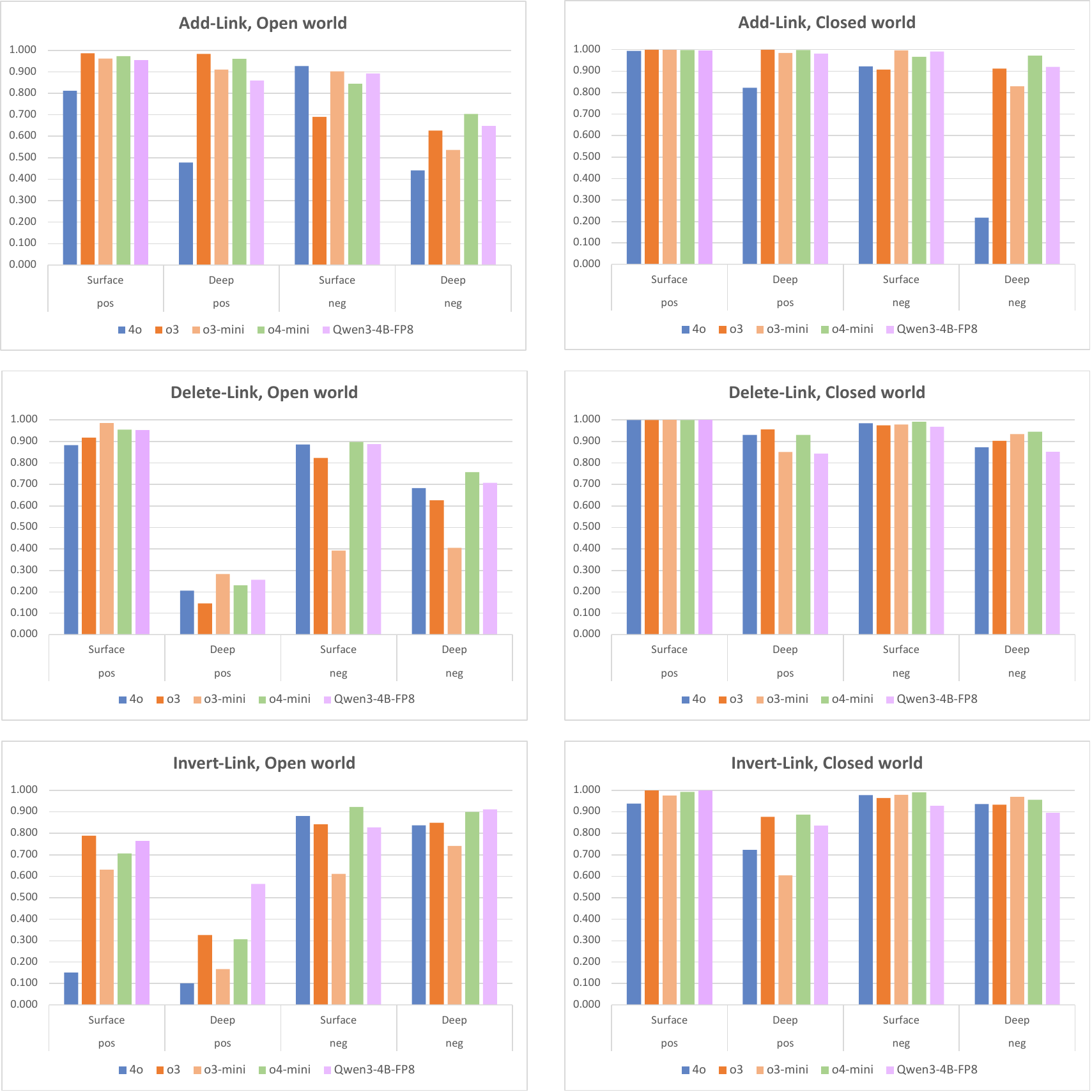}
\caption[Detailed model comparisons]{Comparing models' response accuracy on Counterfactual query types, for the Open and Closed world modes.}
\label{fig:Counterfactuals-model-comparison}
\end{figure}

Accuracy metrics (proportion of samples with correct response) are shown in fig.~\ref{fig:Counterfactuals-model-comparison} and Table~\ref{tab:Counterfactuals-OpenWorld} for the Open World setting, where the model is required to retrieve the relevant MoAs as well as reason about them, and Table~\ref{tab:Counterfactuals-ClosedWorld} for the Closed World setting, where the required MoA (two MoAs for Add-Link) is provided in the prompt. The first column indicates whether the counterfactual link starts at the Drug mentioned in the prompt, i.e. whether it is a \textit{surface} or \textit{deep} counterfactual. The second column distinguishes positive and negative samples. Accuracy for Add-Link queries measures whether an MoA was returned.

General trends can be seen in Table~\ref{tab:Counterfactuals-Groups} and fig.~\ref{fig:Counterfactuals-Groups}, which shows means with confidence intervals, averaged across the grouped cases from Tables~\ref{tab:Counterfactuals-OpenWorld} and \ref{tab:Counterfactuals-ClosedWorld}. The overall performance of the models, in decreasing order, is Qwen3-4B and o4-mini in near ties, followed by o3, o3-mini, and 4o (first two rows in Table~\ref{tab:Counterfactuals-Groups}). While it is not surprising that the reasoning models perform better at the reasoning tasks than the general purpose 4o, and also that o3 performs better than the smaller o3-mini, it is surprising that the smaller o4-mini model outperforms o3, and the small Qwen3-4B model performs so well. There are some cases where Qwen3-4B outperforms o4-mini and others where the situation is reversed. In addition, there are cases where on average, o3-mini outperforms the larger o3 (e.g. closed-world negatives, and surface counterfactuals).

% . . . . . . . . . . . . . . . . . . . . . . . . . . . . . . . . . . . . . . . . . . . . . . . . . . . . . . . . . . . . . . . . . . . . . . . . . . . . . . . .
\subsubsection{MoAs returned for Add-Link queries}

The metrics for directional consistency of the returned MoA to the reference MoA for positive Add-Link samples are shown in Tables~\ref{tab:AddLink-pos-MoA-consistency-ow} (open world) and \ref{tab:AddLink-pos-MoA-consistency-cw} (closed world). The graph match metrics are averages for when an MoA is returned.

In the open world queries, the o3 model returns the most MoAs (measured as `accuracy' in the other tables), with the highest interior node match scores. Not surprisingly, the closed world scores are high for all the metrics, much higher than for open world queries. Closed world queries also return no surprises (`Potentially very different MoAs'), indicating that all the models understand the task, reason correctly most of the time, and when they do return a MoA, it is the expected one. Accuracies (`MoA returned') for deep counterfactuals lags those for surface counterfactuals for both open and closed world queries. More on this below.

% . . . . . . . . . . . . . . . . . . . . . . . . . . . . . . . . . . . . . . . . . . . . . . . . . . . . . . . . . . . . . . . . . . . . . . . . . . . . . . . .
\subsubsection{Open v/s Closed World}

In general, all models' performance in the closed world setting, where all the relevant knowledge is provided, is better than in the open world setting (summarized in rows 1--2, Table~\ref{tab:Counterfactuals-Groups}, and fig.~\ref{fig:Counterfactuals-Groups}). The overall high metrics in the closed world setting demonstrates that the models do understand the language of MoAs sufficiently to reason about them, and also that the prompts we used were understood.

% . . . . . . . . . . . . . . . . . . . . . . . . . . . . . . . . . . . . . . . . . . . . . . . . . . . . . . . . . . . . . . . . . . . . . . . . . . . . . . . .
\subsubsection{Performance on Positives v/s Negatives}
\label{subsec:Pos-Neg}

In the open world setting, model accuracy on the negative samples is much better than on the positive samples (except for o3-mini; rows 3--4, Table~\ref{tab:Counterfactuals-Groups}, and fig.~\ref{fig:Counterfactuals-Groups}). However in the closed world setting, accuracies on the positives and negatives are much closer. Looking at the detailed results for open world queries in Table~\ref{tab:Counterfactuals-OpenWorld}, summarized in rows 7--8 in Table~\ref{tab:Counterfactuals-Groups}, accuracies for positives on Add-Link queries is actually higher than for the corresponding negatives.

The target entity in negative Delete/Invert-Link counterfactuals was randomly selected, so it is unlikely to have been associated in biomedical literature with the query's drug or disease. This probably makes these negatives an easier task. For positives, the model has to retrieve the nature of the association and reason about it. However, both positive and negative Add-Link queries require the same information, so the better performance on positives here is unexpected.

The largest increase in accuracy of negatives over positives is for Delete-Link and Invert-Link deep counterfactuals, summarized in the last two rows of Table~\ref{tab:Counterfactuals-Groups}.
Tabulating the LLM responses for positive queries in this group we observed that `Partially Blocked' was the most commonly returned option (55\% of the responses). This was sometimes justified, as in one example when the model pointed out that the drug had other pathways to the disease that are not blocked by the counterfactual, which we were able to confirm by searching the web. This underscored that DrugMechDB is not a complete representation of all Drug-Disease mechanisms. In other cases, `Partially Blocked' was not justified, e.g. in an Invert-Link counterfactual presented as ``the drug Ceftriaxone increases activity of the Gene Penicillin-binding protein 2 (PBP2)'', where the normal action was for it to decrease that protein's activity. In this example the o4-mini model reasoned that ``Upregulating PBP2 could in theory require higher Ceftriaxone concentrations to achieve the same degree of PBP inhibition'', which is contrary to logic.

If we add `Partially Blocked' as an accepted response, the updated accuracies for Delete and Invert-Link open world positive examples show a significant jump (Table~\ref{tab:Counterfactuals-OpenWorld-Relaxed}). Performance on deep counterfactuals for Invert-Link now matches that for surface counterfactuals, but it still lags for Delete-Link queries. However, as supported by inspecting some examples, the true metrics are somewhere in between the strict and relaxed numbers, and performance on deep counterfactuals would continue to lag that of surface counterfactuals.

Negative counterfactuals for these query types describe an interaction to some randomly selected proteins that have no role in the relevant MoA, so we do not relax the accepted responses for those queries. In fact looking at the justifications the models provide for incorrect responses to these queries often show the model assuming that the mentioned counterfactual is actually a part of the desired MoA. Which also explains why the corresponding closed world metrics are much better.

% . . . . . . . . . . . . . . . . . . . . . . . . . . . . . . . . . . . . . . . . . . . . . . . . . . . . . . . . . . . . . . . . . . . . . . . . . . . . . . . .
\subsubsection{Deep Counterfactuals}

\begin{figure}
\centering
\includegraphics[width=\columnwidth]{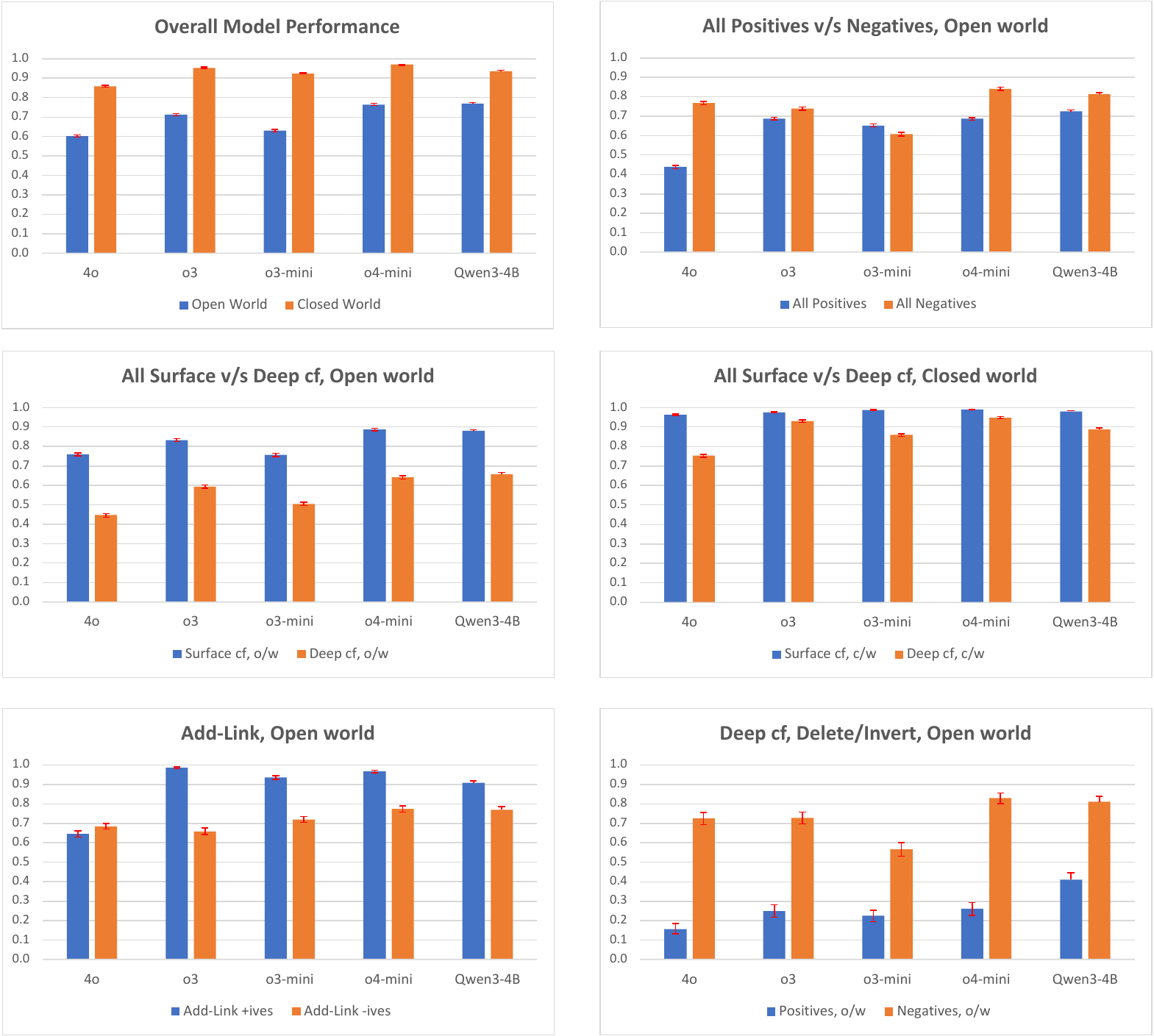}
\caption[Charts for the grouped accuracies]{Charts for some of the grouped accuracies for Counterfactuals, showing trends.}
\label{fig:Counterfactuals-Groups}
\end{figure}

Rows 9--12 in Table~\ref{tab:Counterfactuals-Groups}, and fig.~\ref{fig:Counterfactuals-Groups}, show that deep counterfactuals are a harder task for all four models, in both open and closed world settings. This is probably because in surface counterfactuals, the `observed' counterfactual relationship is to the drug mentioned in the prompt, whereas in the deep examples the connection of the counterfactual's entities to the drug and disease mentioned in the prompt is indirect and requires recall (open world) and inference. In the closed world examples, models o3 and o4-mini show the least difference in performance between surface and deep counterfactuals.

The cases with the worst performance across all models are open world positive samples for when an interior link is either deleted or inverted (`deep $+$ive', rows 6 and 10 in Table~\ref{tab:Counterfactuals-OpenWorld}). Closed world performance for these cases is quite a bit better, although for positive Invert-Link samples (row 10 in Table~\ref{tab:Counterfactuals-ClosedWorld}), performance on deep counterfactuals still significantly lags that for surface counterfactuals.

\begin{table}
% src: Sessions/Latest/Qwen3-4B-Thinking-2507-FP8/AddLink_pos_{d,p}pi_r1k_log.txt
\renewcommand{\arraystretch}{1.3}
\centering
\caption{MoA consistency metrics for Add-Link positives, Open World.}
\label{tab:AddLink-pos-MoA-consistency-ow}
%\resizebox{\columnwidth}{!}{%
\begin{tabular}{lrrrrr}
\hline
\bfseries Metric & \bfseries 4o & \bfseries o3 & \bfseries o3-mini & \bfseries o4-mini & \bfseries Qwen3-4B \\
\hline
\multicolumn{6}{l}{\hspace{2em}\textit{surface counterfactuals} \ldots} \\
MoA returned  &  0.812  &  0.987  &  0.962  &  0.973 & 0.955 \\
Potentially v. different MoAs  &  15.9\%  &  20.9\%  &  25.5\%  &  25.8\% & 28.5\%\\
avg. Interior Node-match score  &  0.513  &  0.545  &  0.490  &  0.510 & 0.470 \\
avg. Reduced Edge-match score  &  0.989  &  0.984  &  0.982  &  0.967 & 0.968 \\
\multicolumn{6}{l}{\hspace{2em}\textit{deep counterfactuals} \ldots} \\
MoA returned  &  0.478  &  0.984  &  0.911  &  0.961 & 0.860 \\
Potentially v. different MoAs  &  7.9\%  &  20.8\%  &  29.1\%  &  27.7\% & 25.7\% \\
avg. Interior Node-match score  &  0.608  &  0.627  &  0.559  &  0.584 & 0.537 \\
avg. Reduced Edge-match score  &  0.990  &  0.992  &  0.987  &  0.985 & 0.973 \\
\hline
\end{tabular}
%}
\end{table}

\begin{table}
% src: Sessions/Latest/Qwen3-4B-Thinking-2507-FP8/AddLink_pos_{d,p}pi_r1k-k_log.txt
\renewcommand{\arraystretch}{1.3}
\centering
\caption{MoA consistency metrics for Add-Link positives, Closed World.}
\label{tab:AddLink-pos-MoA-consistency-cw}
%\resizebox{\columnwidth}{!}{%
\begin{tabular}{lrrrrr}
\hline
\bfseries Metric & \bfseries 4o & \bfseries o3 & \bfseries o3-mini & \bfseries o4-mini & \bfseries Qwen3-4B \\
\hline
\multicolumn{6}{l}{\hspace{2em}\textit{surface counterfactuals} \ldots} \\
MoA returned  &  0.995  &  1.000  &  1.000  &  0.999 & 0.997 \\
Potentially v. different MoAs  &  0\%  &  0\%  &  0\%  &  0\% & 0\% \\
avg. Interior Node-match score  &  0.995  &  0.991  &  0.974  &  0.998 & 0.970 \\
avg. Reduced Edge-match score  &  0.999  &  0.997  &  0.995  &  0.999 & 0.992 \\
\multicolumn{6}{l}{\hspace{2em}\textit{deep counterfactuals} \ldots} \\
MoA returned  &  0.822  &  1.000  &  0.985  &  0.999 & 0.982 \\
Potentially v. different MoAs  &  0\%  &  0\%  &  0\%  &  0\% & 0\% \\
avg. Interior Node-match score  &  0.988  &  0.983  &  0.933  &  0.996 & 0.949 \\
avg. Reduced Edge-match score  &  0.999  &  0.998  &  0.991  &  1.000 & 0.991 \\
\hline
\end{tabular}
%}
\end{table}

\begin{table}
% src: Sessions/Latest/Variances/Summary_edit.{md,xlsx}; Qwen: Sessions/Latest/cfmetrics.md
\renewcommand{\arraystretch}{1.3}
\centering
\caption{Accuracy results for Counterfactuals, Open World setting.}
\label{tab:Counterfactuals-OpenWorld}
\begin{tabular}{ccrrrrr}
\hline
\bfseries cf. sub-type & \bfseries Pos/Neg & \bfseries 4o & \bfseries o3 & \bfseries o3-mini & \bfseries o4-mini & \bfseries Qwen3-4B \\
\hline
\multicolumn{7}{l}{\hspace{0em}\textit{Add-Link} \ldots} \\
surface & +ive & 0.812 & 0.987 & 0.962 & 0.973 & 0.955 \\
deep    & +ive & 0.478 & 0.984 & 0.911 & 0.961 & 0.860 \\
surface & $-$ive & 0.927 & 0.690 & 0.903 & 0.844 & 0.892 \\
deep    & $-$ive & 0.441 & 0.627 & 0.536 & 0.704 & 0.648 \\
\multicolumn{7}{l}{\hspace{0em}\textit{Delete-Link} \ldots} \\
surface & +ive & 0.883 & 0.918 & 0.985 & 0.955 & 0.952 \\
deep    & +ive & 0.206 & 0.146 & 0.283 & 0.230 & 0.256 \\
surface & $-$ive & 0.886 & 0.824 & 0.392 & 0.898 & 0.888 \\
deep    & $-$ive & 0.683 & 0.626 & 0.404 & 0.758 & 0.708 \\
\multicolumn{7}{l}{\hspace{0em}\textit{Invert-Link} \ldots} \\
surface & +ive & 0.151 & 0.788 & 0.632 & 0.706 & 0.764 \\
deep    & +ive & 0.101 & 0.326 & 0.167 & 0.306 & 0.564 \\
surface & $-$ive & 0.881 & 0.842 & 0.611 & 0.922 & 0.828 \\
deep    & $-$ive & 0.837 & 0.850 & 0.741 & 0.900 & 0.912 \\
\hline
\end{tabular}
\end{table}

\begin{table}
% src: Sessions/Latest/Variances/Summary_edit.{md,xlsx}; Qwen: Sessions/Latest/cfmetrics.md
\renewcommand{\arraystretch}{1.3}
\centering
\caption{Accuracy results for Counterfactuals, Closed World.}
\label{tab:Counterfactuals-ClosedWorld}
\begin{tabular}{ccrrrrr}
\hline
\bfseries cf. sub-type & \bfseries Pos/Neg & \bfseries 4o & \bfseries o3 & \bfseries o3-mini & \bfseries o4-mini & \bfseries Qwen3-4B \\
\hline
\multicolumn{7}{l}{\hspace{0em}\textit{Add-Link} \ldots} \\
surface & +ive & 0.995 & 1.000 & 1.000 & 0.999 & 0.987 \\
deep    & +ive & 0.822 & 1.000 & 0.985 & 0.999 & 0.982 \\
surface & $-$ive & 0.923 & 0.908 & 0.997 & 0.967 & 0.992 \\
deep    & $-$ive & 0.217 & 0.912 & 0.829 & 0.973 & 0.920 \\
\multicolumn{7}{l}{\hspace{0em}\textit{Delete-Link} \ldots} \\
surface & +ive & 1.000 & 1.000 & 1.000 & 0.999 & 1.000 \\
deep    & +ive & 0.930 & 0.957 & 0.850 & 0.930 & 0.844 \\
surface & $-$ive & 0.985 & 0.975 & 0.979 & 0.992 & 0.968 \\
deep    & $-$ive & 0.873 & 0.903 & 0.934 & 0.946 & 0.852 \\
\multicolumn{7}{l}{\hspace{0em}\textit{Invert-Link} \ldots} \\
surface & +ive & 0.938 & 1.000 & 0.977 & 0.993 & 1.000 \\
deep    & +ive & 0.722 & 0.876 & 0.605 & 0.888 & 0.836 \\
surface & $-$ive & 0.978 & 0.965 & 0.980 & 0.990 & 0.928 \\
deep    & $-$ive & 0.936 & 0.933 & 0.970 & 0.957 & 0.896 \\
\hline
\end{tabular}
\end{table}

\begin{table}
% src: Sessions/Latest/LatestMetrics.xlsx; Qwen: Sessions/Latest/cf_grouped_metrics.md
\renewcommand{\arraystretch}{1.3}
\centering
\caption{Grouped accuracies for Counterfactuals, showing trends.}
\label{tab:Counterfactuals-Groups}
%\resizebox{\columnwidth}{!}{%
\begin{tabular}{lrrrrr}
\hline
\bfseries Group & \bfseries 4o & \bfseries o3 & \bfseries o3-mini & \bfseries o4-mini & \bfseries Qwen3-4B \\
\hline
All, o/w   &  0.60$\,\pm\,$0.01  &  0.71$\,\pm\,$0.01  &  0.63$\,\pm\,$0.01  &  0.76$\,\pm\,$0.01  &  0.77$\,\pm\,$0.01 \\
All, c/w &  0.86$\,\pm\,$0.00  &  0.95$\,\pm\,$0.00  &  0.92$\,\pm\,$0.00  &  0.97$\,\pm\,$0.00  &  0.94$\,\pm\,$0.00\\
\hline
Positives, o/w   &  0.44$\,\pm\,$0.01  &  0.69$\,\pm\,$0.01  &  0.65$\,\pm\,$0.01  &  0.69$\,\pm\,$0.01 &  0.73$\,\pm\,$0.01  \\
Negatives, o/w &  0.77$\,\pm\,$0.01  &  0.74$\,\pm\,$0.01  &  0.61$\,\pm\,$0.01  &  0.84$\,\pm\,$0.01  &  0.81$\,\pm\,$0.01 \\
\hline
Positives, c/w   &  0.90$\,\pm\,$0.01  &  0.97$\,\pm\,$0.00  &  0.90$\,\pm\,$0.01  &  0.97$\,\pm\,$0.00  &  0.94$\,\pm\,$0.01 \\
Negatives, c/w &  0.82$\,\pm\,$0.01  &  0.93$\,\pm\,$0.01  &  0.95$\,\pm\,$0.01  &  0.97$\,\pm\,$0.00  &  0.91$\,\pm\,$0.01 \\
\hline
AddLink +ive, o/w  &  0.65$\,\pm\,$0.02  &  0.99$\,\pm\,$0.01  &  0.94$\,\pm\,$0.01  &  0.97$\,\pm\,$0.01 &  0.91$\,\pm\,$0.01  \\
AddLink -ive, o/w  &  0.68$\,\pm\,$0.01  &  0.66$\,\pm\,$0.02  &  0.72$\,\pm\,$0.01  &  0.77$\,\pm\,$0.01 &  0.77$\,\pm\,$0.01 \\
\hline
Surface cf, o/w  &  0.76$\,\pm\,$0.01  &  0.83$\,\pm\,$0.01  &  0.76$\,\pm\,$0.01  &  0.89$\,\pm\,$0.01 &  0.88$\,\pm\,$0.01 \\
Deep cf, o/w     &  0.45$\,\pm\,$0.01  &  0.59$\,\pm\,$0.01  &  0.51$\,\pm\,$0.01  &  0.64$\,\pm\,$0.01 &  0.66$\,\pm\,$0.01 \\
\hline
Surface cf, c/w  &  0.96$\,\pm\,$0.00  &  0.97$\,\pm\,$0.00  &  0.99$\,\pm\,$0.00  &  0.99$\,\pm\,$0.00 &  0.98$\,\pm\,$0.00  \\
Deep cf, c/w     &  0.75$\,\pm\,$0.01  &  0.93$\,\pm\,$0.01  &  0.86$\,\pm\,$0.01  &  0.95$\,\pm\,$0.01 &  0.89$\,\pm\,$0.01  \\
\hline
D.D.I. pos, o/w   &  0.16$\,\pm\,$0.03  &  0.25$\,\pm\,$0.03  &  0.22$\,\pm\,$0.03  &  0.26$\,\pm\,$0.03 &  0.41$\,\pm\,$0.04 \\
D.D.I. neg, o/w   &  0.73$\,\pm\,$0.03  &  0.73$\,\pm\,$0.03  &  0.57$\,\pm\,$0.03  &  0.83$\,\pm\,$0.03 &  0.81$\,\pm\,$0.03 \\
\hline
\multicolumn{5}{c}{Mean group accuracy $\pm$ 90\% c.i. (rounded) derived using stratified bootstrap.}\\
\multicolumn{5}{c}{o/w = Open World, c/w = Closed World, cf = counterfactuals}\\
\multicolumn{5}{c}{D.D.I. = Deep cf for Delete / Invert queries}\\
\end{tabular}
%}
\end{table}

\begin{figure}
\centering
\subfloat[Depth from root]{\includegraphics[width=0.5\columnwidth]{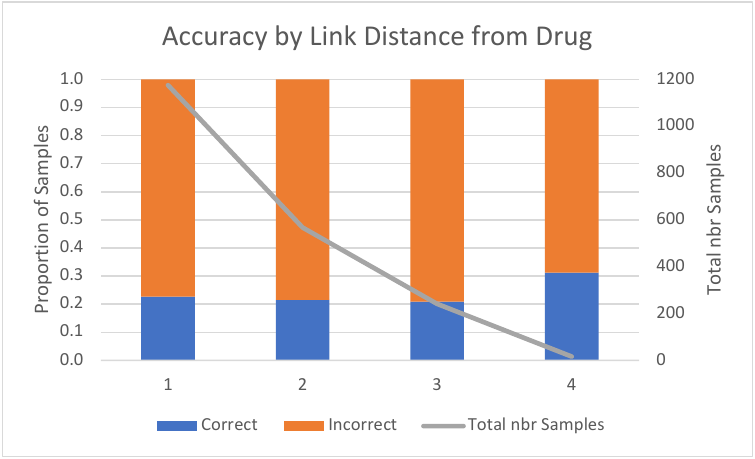}
\label{fig:pos_ppi_dist_to_root}
}
\hfil
\subfloat[Distance to sink]{\includegraphics[width=0.5\columnwidth]{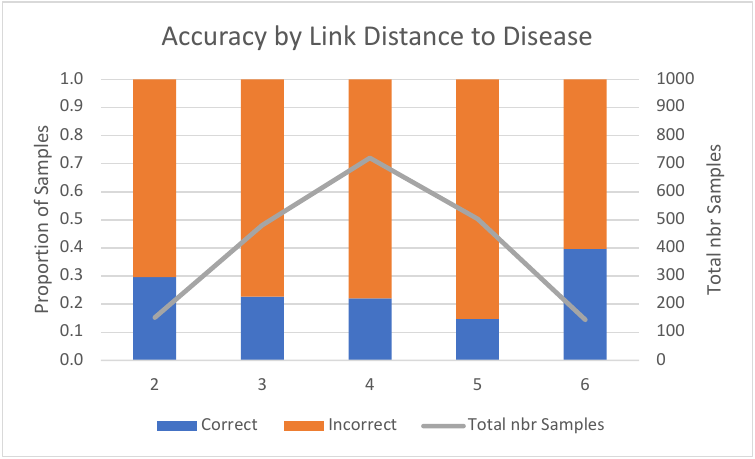}
\label{fig:pos_ppi_dist_to_sink}
}
\hfil
\subfloat[MoA length]{\includegraphics[width=0.5\columnwidth]{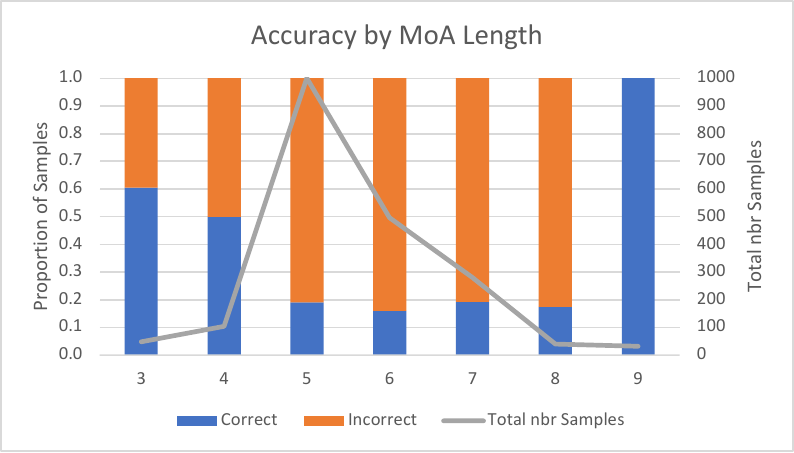}
\label{fig:pos_ppi_total_link_dist}
}
\caption{Exploring the effect of longer reasoning chains on deep counterfactuals; positive samples, open world, OpenAI models.}
\label{fig:deep_reasoning_length}
\end{figure}

We investigated whether the length of the reasoning chains represented by the MoAs, and the depth of the counterfactuals in this inference chain, affected the performance for deep counterfactuals (fig.~\ref{fig:deep_reasoning_length}). The data for these charts combines the open world positive samples for deep counterfactuals, for the Delete and Invert-Link query types, across the four OpenAI models (second-last row in Table~\ref{tab:Counterfactuals-Groups}): 2,000 instances of which 444 (22.2\%) received a correct response and 1,556 (77.8\%) an incorrect response. The first two charts plot model accuracy (proportion of correct v/s incorrect responses) by depth of the counterfactual's source node from the root or drug (fig.~\ref{fig:pos_ppi_dist_to_root}), and distance of the counterfactual's source node from the sink or disease (fig.~\ref{fig:pos_ppi_dist_to_sink}). We also examined the overall length of the MoA (shortest path passing through the counterfactual's source node, fig.~\ref{fig:pos_ppi_total_link_dist}). The charts suggest a downward trend of accuracy with increasing MoA length and possibly distance from the disease node. However, a logistic regression model trained to predict the LLM response given these three distances as variables does not yield significant co-efficients other than for the intercept. The `noise' in response validation discussed above (\S~\ref{subsec:Pos-Neg}) may be a confounding factor here. In conclusion, deep counterfactuals do present a problem, and how this affects LLMs' reasoning capabilities is worth investigating further.

\begin{table}
\renewcommand{\arraystretch}{1.3}
\centering
\caption{Relaxed Accuracy results for selected positive Counterfactuals, Open World setting.}
\label{tab:Counterfactuals-OpenWorld-Relaxed}
\begin{tabular}{ccrrrrr}
\hline
\bfseries cf. sub-type & \bfseries Pos/Neg & \bfseries 4o & \bfseries o3 & \bfseries o3-mini & \bfseries o4-mini & \bfseries Qwen3-4B \\
\hline
\multicolumn{7}{l}{\hspace{0em}\textit{Delete-Link} \ldots} \\
surface & +ive & 0.950 & 0.978 & 0.993 & 0.978 & 0.964 \\
deep    & +ive & 0.849 & 0.850 & 0.853 & 0.829 & 0.640 \\
\multicolumn{7}{l}{\hspace{0em}\textit{Invert-Link} \ldots} \\
surface & +ive & 0.472 & 0.848 & 0.706 & 0.747 & 0.804 \\
deep    & +ive & 0.549 & 0.853 & 0.607 & 0.781 & 0.584 \\
\hline
\end{tabular}
\end{table}

\section{Related Work}
\label{sec:RelatedWork}

% --------------------------------------------------------------------------------------------------------------------------------------
\subsection{Biomedical Q\&A Datasets}
MultiMedQA \citep{Singhal-etal:2022:MultiMedQA} is a curated compilation of multiple choice questions derived from six previously published medical datasets and one new dataset on general medical knowledge. Most questions are taken from high school to professional exams available online, or derived from other online publications, e.g. PubMedQA \citep{Jin-etal:2019:PubMedQA} has questions artificially generated from PubMed abstracts.

LLMs fine-tuned on the medical domain (e.g. Med PaLM 2 \citep{Singhal-etal:2025:MedPaLM-2}) have achieved expert level performance on such datasets. However the work on MedPrompt \citep{Nori-etal:2023:Medprompt} demonstrates that ensembling, and using a composition of specialized prompt techniques, including selecting few-shot examples tailored to each query, can push the general purpose GPT-4 to surpass these results. Focusing primarily on the development of our counterfactuals dataset, we have kept our queries simple.

Noting that ``models that perform well on these benchmarks can still struggle in scenarios requiring fine-grained clinical reasoning'', \citep{Tagliabue-etal:2025:DeVisE} manually constructed the DeVisE benchmark for clinical NLP based on MIMIC-IV discharge summaries. They included counterfactual questions in which one clinical variable, a demographic or a vital sign, was altered. Counterfactual queries were generated using automated techniques, and manually reviewed. Our counterfactuals deal with molecular interactions, and were also automatically derived using graph techniques and an MoA simulator to ensure correctness.

The Neuropathic pain dataset \citep{Tu-etal:2019} is a small causal graph describing interactions in neuropathic pathophysiology, used in \citep{Kiciman-etal:2024} to evaluate LLMs on their ability to predict the correct causal relationship between biomedical entities.

One of the tasks (`pharmacology') in the DrugPC dataset introduced in \citep{Gao-etal:2025:TxAgent} contains a small number of queries where the model has to consider the queried drug's mechanism of action. They also do not explore counterfactual and negative variants as exhaustively.
Also motivated by \citep{Gallifant-etal:2024}, the DescriptionPC dataset in \citep{Gao-etal:2025:TxAgent} anonymizes the drug, replacing it with a detailed description that includes indications, mechanisms of action, contraindications, and interactions. We tested anonymizing the disease in our factual knowledge test.

% --------------------------------------------------------------------------------------------------------------------------------------
\subsection{Reasoning}
There are several recent surveys of the evaluation of reasoning capabilities of general purpose pre-trained LLMs \citep{Sun-etal:2024,Yu-etal:2024,Mondorf-etal:2024}. Reasoning can be informally defined as ``\textit{the process of making inferences based on existing knowledge}'' \citep{Yu-etal:2024}. A Mechanism of Action graph can be viewed as an enhanced causal diagram \citep{Pearl-Mackenzie:2018:Book} that also includes physical relationships, and the causal edges are labeled with their direction-of-effect. More formally, asking for a MoA to support a Drug-\textit{treats}-Disease assertion is a form of \textit{Abductive Reasoning} (Wikipedia\footnote{\url{https://en.wikipedia.org/wiki/Abductive_reasoning}} and \citep{Sun-etal:2024,Yu-etal:2024}). Reference \citep{Sun-etal:2024} also has a good review of research investigating reasoning in the Medical and Bioinformatics domains, and \citep{Yang-etal:2024:CausalBenchmarks} reviews causal reasoning benchmarks for LLMs.

% --------------------------------------------------------------------------------------------------------------------------------------
\subsection{Counterfactuals}
While counterfactuals are an integral part of causal reasoning, interest in counterfactual-based queries for evaluating LLMs on reasoning tasks was also motivated by the need to measure generalizable reasoning capabilities, rather than models' ability to simply answer problems similar to those seen during training (e.g. \citep{Wu-etal:2024}).

Early work on evaluating LLM reasoning capability using counterfactuals, e.g. \citep{Qin-etal:2019, Yang-etal:2020:SemEval, Frohberg-Binder:2022:CRASS}, presented altered events or situations that conformed to the `world model' encountered by LLMs in training text. In contrast, \citep{Wu-etal:2024} was perhaps the first work to test LLMs on counterfactuals that described an alternative world model with different conditions (e.g. chess start positions, or arithmetic in base 11), while still keeping the basic reasoning rules. Our counterfactuals describe a change from how biomedical entities (e.g. drugs, proteins) are known to interact, which then affects the mechanistic viability of some drugs for specific diseases. This is closer in principle to the alternative world model of \citep{Wu-etal:2024}.

Each query in the CRASS dataset \citep{Frohberg-Binder:2022:CRASS} contains the base fact, the counterfactualized query, and the optional conclusions, and relies on a `common sense' world and reasoning model. This is similar to our closed world setting; in our open world setting, we only provide the counterfactualized query, and the LLM is expected to recall all knowledge, including other interactions, that are relevant to the task. We believe this setting better reflects how LLM-based tools will be used to assist scientific research.

CRASS was retested in \citep{Kiciman-etal:2024}, finding that the newer GPT-4 outperformed GPT-3.5. Our results show that this progression in performance has continued with newer OpenAI models.

Counterfactuals have also been used for other purposes, e.g. to evaluate social biases in LLMs for medical applications \citep{Poulain-etal:2024}; exploring interventions in patient treatment and their resulting patient trajectories or the effects of genetic perturbations \citep{Li-etal:2025:CLEF}; to improve an LLM's reasoning ability by including counterfactual questions in its fine-tuning \citep{Huyuk-etal:2025}.

\section{Conclusion}
\label{sec:Conclusion}

We have introduced the first large dataset, containing positive and negative counterfactuals, for evaluating an LLM's knowledge and understanding of drug and disease mechanisms, with a special focus on the LLMs' ability to reason in this domain. Using this dataset to evaluate five large language models reveals some interesting behavior patterns: (i) Most models perform better on negative examples than on positive examples, possibly because they offer stronger syntactic cues; o3-mini is the one exception showing very similar performance for both sets. (ii) Deep counterfactuals present a much harder task than surface counterfactuals for all models.

The closed world experiments demonstrate that these models do understand the language of biomedical interactions sufficiently to be able to reason about drug-disease mechanisms. However the more realistic open world experiments reveal that combining recall of relevant knowledge with reasoning is a harder task. As expected, the reasoning models show better performance than 4o, with the newest small reasoning models Qwen3-4B and o4-mini performing the best.

Biology and Medicine are constantly evolving, and we continue to get deeper understanding of disease (e.g. COVID) and drug (e.g. GLP-1 RAs) mechanisms. So LLM responses in such critical and deep technical fields should always be validated by an expert. When used as a platform for tools in scientific research, it will be important to combine pre-trained LLMs with an actively updated knowledge base, to keep its factual knowledge current.

%%==============================================================================================

%\clearpage

% vertical separation between references
\setlength\bibsep{5pt}

\bibliography{langkg}
\bibliographystyle{plainnat}

%
% Use \makesignature command to print the information about the
% authors.
%
\vspace{0.4in}
\makesignature

\end{document}